\documentclass[11pt,a4paper]{article}
\usepackage[hyperref]{naaclhlt2019}
\usepackage{times}
\usepackage{latexsym}
\usepackage{graphicx}
\usepackage{color}
\usepackage{url}
\usepackage{multirow}
\usepackage{amsmath}
\usepackage{amssymb}
\usepackage{tabularx}
\usepackage{multicol}
\usepackage{subcaption}

\newcommand{\shorteq}{%
 \rule[.4ex]{4pt}{0.4pt}\llap{\rule[.7ex]{4pt}{0.4pt}}}

\aclfinalcopy 
\title{Handling Noisy Labels for Robustly Learning from Self-Training Data for Low-Resource Sequence Labeling}

\author{Debjit Paul$^*$\thanks{This work was started while the authors were at Saarland University.}, Mittul Singh$^{\dagger\S}$, Michael A. Hedderich$^\ddagger$, Dietrich Klakow$^\ddagger$
 \\ $^*$Research Training Group AIPHES, Institute for Computational Linguistics,\\ Heidelberg University, Germany\\
 $^\dagger$Department of Signal Processing and Acoustics, Aalto University, Finland\\
 $^\ddagger$Spoken Language Systems (LSV), Saarland Informatics Campus,\\ Saarland University, Germany\\
 \small \tt paul@cl.uni-heidelberg.de, mittul.singh@aalto.fi, \\
 \small \tt \{mhedderich, dietrich.klakow\}@lsv.uni-saarland.de
}

\date{}
\begin{document}
\setlength{\abovedisplayskip}{0pt}
\setlength{\belowdisplayskip}{4pt}
\maketitle
\begin{abstract}
In this paper, we address the problem of effectively self-training neural networks in a low-resource setting. Self-training is frequently used to automatically increase the amount of training data. However, in a low-resource scenario, it is less effective due to unreliable annotations created using self-labeling of unlabeled data. We propose to combine self-training with noise handling on the self-labeled data. Directly estimating noise on the combined clean training set and self-labeled data can lead to corruption of the clean data and hence, performs worse. Thus, we propose the Clean and Noisy Label Neural Network which trains on clean and noisy self-labeled data simultaneously by explicitly modelling clean and noisy labels separately. In our experiments on Chunking and NER, this approach performs more robustly than the baselines. Complementary to this explicit approach, noise can also be handled implicitly with the help of an auxiliary learning task. To such a complementary approach, our method is more beneficial than other baseline methods and together provides the best performance overall.
\end{abstract}
\section{Introduction}
For many low-resource languages or domains, only small amounts of labeled data exist. Raw or unlabeled data, on the other hand, is usually available even in these scenarios. Automatic annotation or distant supervision techniques are an option to obtain labels for this raw data, but they often require additional external resources like human-generated lexica which might not be available in a low-resource context. Self-training is a popular technique to automatically label additional text. There, a classifier is trained on a small amount of labeled data and then used to obtain labels for unlabeled instances. However, this can lead to unreliable or noisy labels on the additional data which impede the learning process \citep{Pechenizkiy1647654,Nettleton2010}. In this paper, we focus on overcoming this slowdown of self-training. Hence, we propose to apply noise-reduction techniques during self-training to clean the self-labeled data and learn effectively in a low-resource scenario.\\
\indent Inspired by the improvements shown by the Noisy Label Neural Network (\textit{NLNN}, \citet{7472164}), we can directly apply \textit{NLNN} to the combined set of the existing clean data and the noisy self-labeled data. However, such an application can be detrimental to the learning process (Section \ref{sec:results}). Thus, we introduce the Clean and Noisy Label Neural Network (\textit{CNLNN}) that treats the clean and noisy data separately while training on them simultaneously (Section \ref{sec:cnlnn}).\\
\indent This approach leads to two advantages over \textit{NLNN} (Section \ref{sec:results} and \ref{sec:analysis}) when evaluating on two sequence-labeling tasks, Chunking and Named Entity Recognition. \textbf{Firstly}, when adding noisy data, \textit{CNLNN} is robust showing consistent improvements over the regular neural network, whereas \textit{NLNN} can lead to degradation in performance. \textbf{Secondly}, when combining with an indirect-noise handling technique, i.e. with an auxiliary target in a multi-task fashion, \textit{CNLNN} complements better than \textit{NLNN} in the multi-task setup and overall leads to the best performance.

\section{Related Work}
Self-training has been applied to various NLP tasks, e.g. \citet{Steedman:2003:BSP:1067807.1067851} and \citet{sagae2007dependency}. While \citet{mcclosky2006effective} are able to leverage self-training for parsing, \citet{Charniak:1997:SPC:1867406.1867499} and \citet{Clark:2003:BPT:1119176.1119183} obtain only minimal improvements at best on parsing and POS-tagging respectively. In some cases, the results even deteriorate. Other successful approaches of automatically labeling data include using a different classifier trained on out-of-domain data \cite{Petrov:2010:UAD:1870658.1870727} or leveraging external knowledge \cite{dembowski2017ner}.\\ 
\indent A detailed review of learning in the presence of noisy labels is given in \cite{6685834}. Recently, several approaches have been proposed for modeling the noise using a confusion matrix in a neural network context. Many works assume that all the data is noisy-labeled \cite{7472164, Goldberger2017NoiseAdaptation,Sukhbaatar2015Learning}. \citet{W18-3402} and \citet{Hendrycks2018Trusted} propose a setting where a mix of clean and unlabeled data is used. However, they require external knowledge sources for labeling the data or evaluate on synthetic noise. Alternatively, instances with incorrect labels might be filtered out, e.g. in the work by \citet{guan2011Identifying} or \citet{Han2018CoteachingRT}, but this involves the risk of also filtering out difficult but correct instances. Another orthogonal approach is the use of noise-robust loss functions \cite{Zhang2018GeneralizedCE}.

\section{Clean and Noisy Label Neural Network}
\label{sec:cnlnn}
\begin{figure}
  \centering
    \includegraphics[scale=0.6]{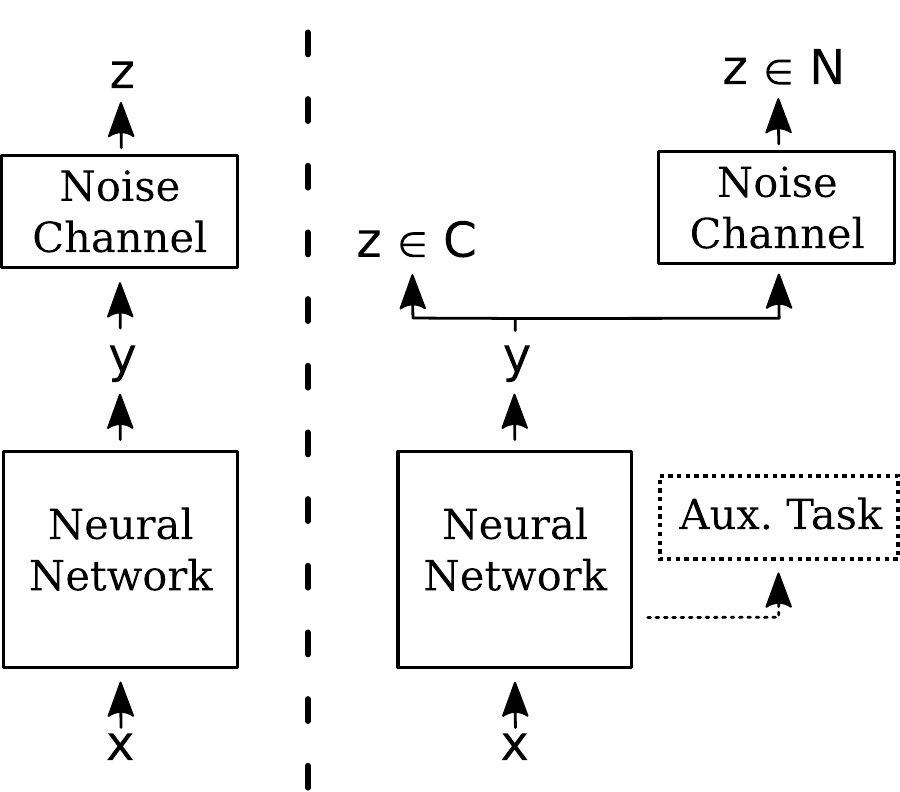}
    \caption{A representation of \textit{NLNN} (left) compared to our proposed \textit{CNLNN} model. The complementary multi-task component (aux. task) is dashed.}
    \label{fig:cnlnn}
\end{figure}

The Noisy Label Neural Network (\textit{NLNN}, \citet{7472164}) assumes that all observed labels in the training set pass through a noise channel flipping some of them from a correct to an incorrect label (see left part of Figure \ref{fig:cnlnn}). In our scenario, this means that both the human-annotated and the additional automatically-labeled (self-training) corpora are assumed to be noisy. In our experiments (Section \ref{sec:results} and \ref{sec:analysis}), treating both corpora in this fashion degrades the overall performance. To remedy this effect, we propose to treat the human-annotated data as clean data and the self-training data as noisy.\\
\indent We assume a similar setup as \citet{7472164}, training a multi-class neural network soft-max classifier 

{\small
\begin{equation*}
 p(y=i|x;w) = \frac{exp(u^T_i h)}{\sum^k_{j=1} exp(u^T_jh)}
\end{equation*}
}
where $x$ is the feature vector, $y$ is the label, $w$ denotes the network weights, $k$ is the number of possible labels, $u$ are soft-max weights and $h=h(x)$ denotes the multi-layer neural network applied to $x$.
In contrast to \newcite{7472164}, we assume that not all of the training data passes through a noisy channel changing the correct labels $y$ to noisy ones ($z \in N$). A part of the training set remains clean ($z \in C$) such that $|C| + |N|~\shorteq~n$ where $n$ is the total number of training examples. The clean labels are a copy of the corresponding correct labels. A schematic representation of this model is shown on the right side of Figure \ref{fig:cnlnn}. The correct labels $y$ and the noise distribution $\theta$ are hidden for the noisy labels.\\ 
\begin{figure*}[!ht]
  \centering
  \begin{subfigure}{0.5\textwidth}
    \includegraphics[width=0.98\linewidth,height=4.9cm]{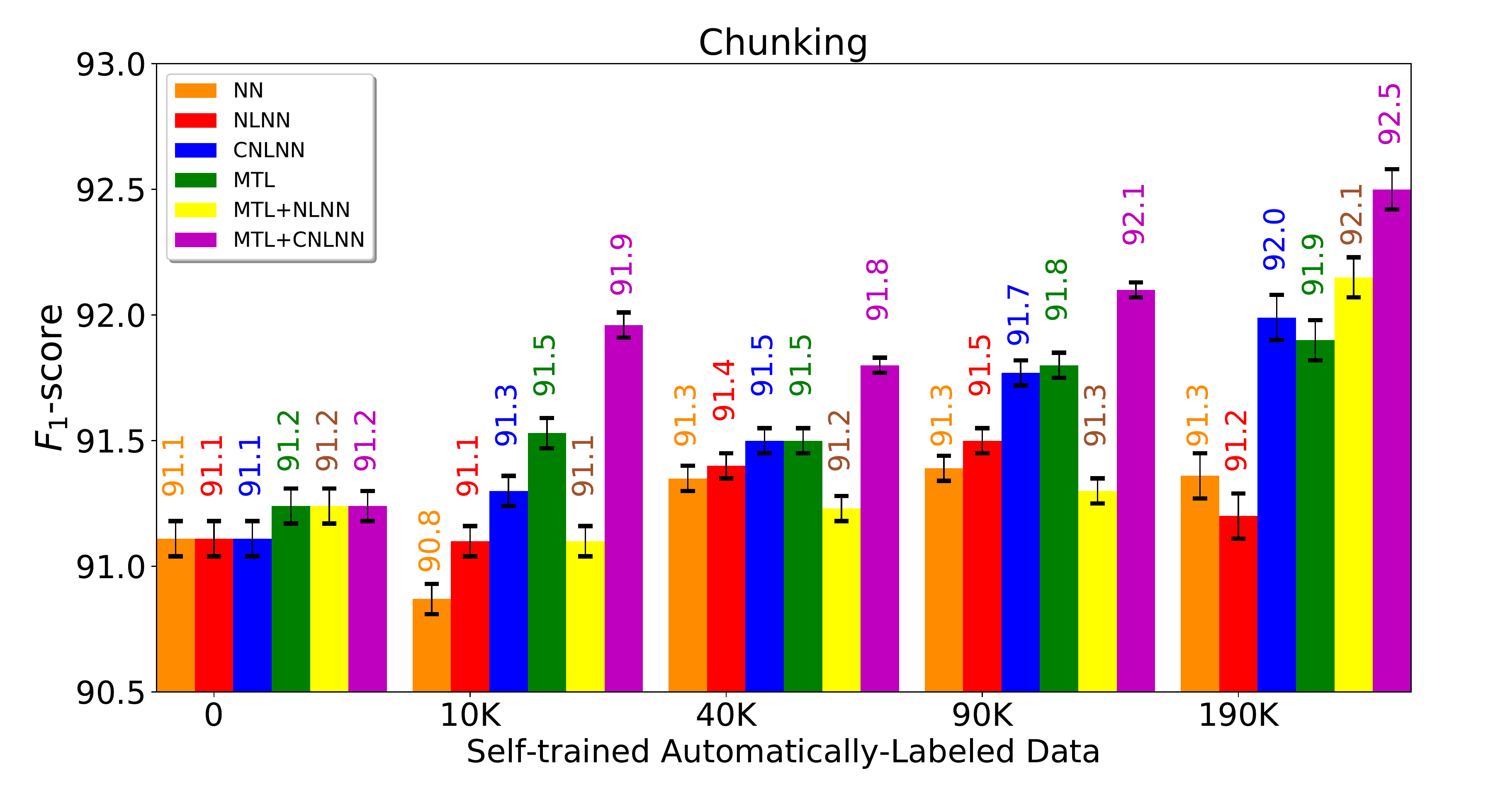}
  \end{subfigure}%
  \begin{subfigure}{0.5\textwidth}
    \includegraphics[width=0.98\linewidth,height=4.9cm]{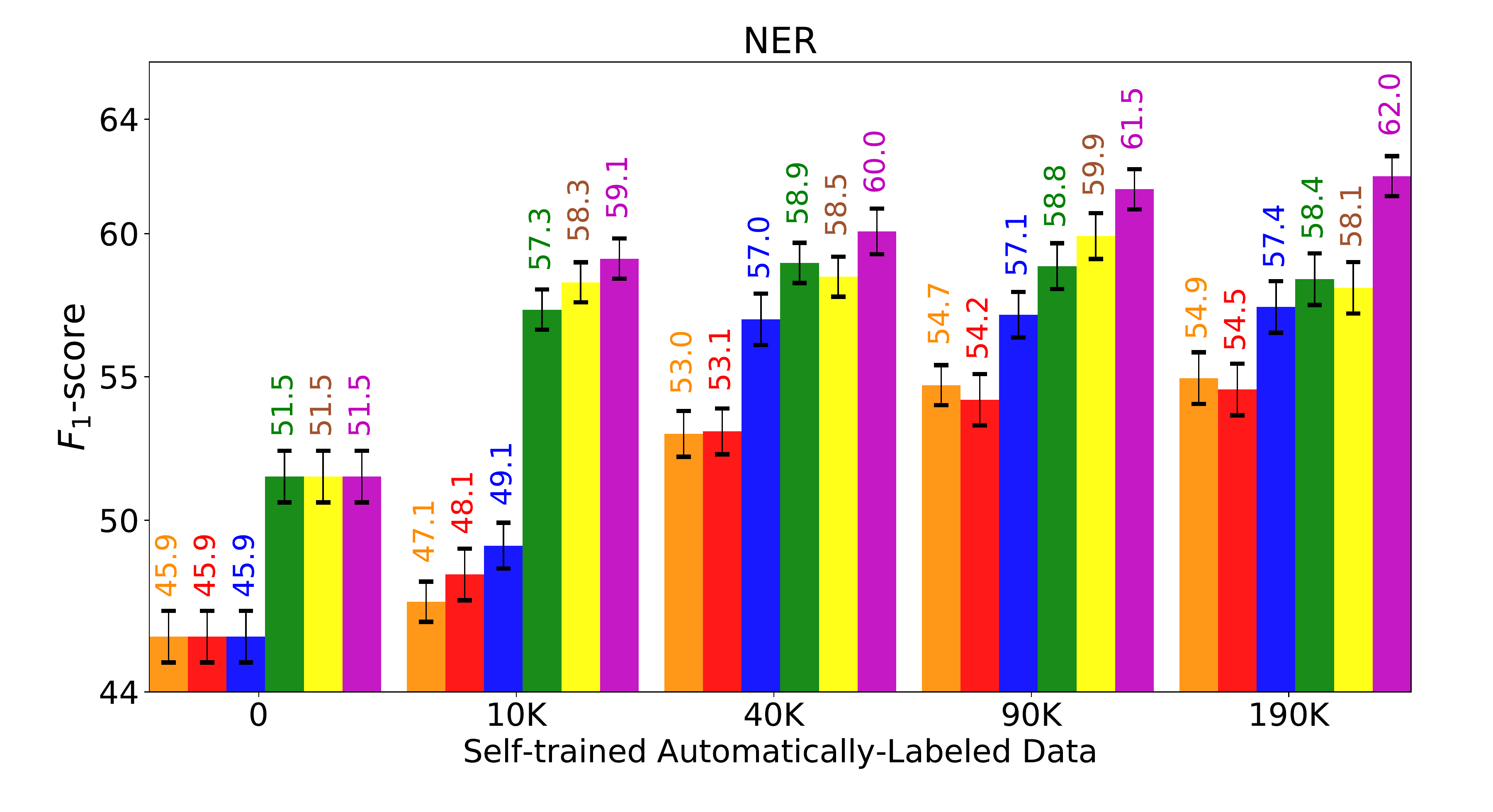}
  \end{subfigure}
    \caption{\label{fig:adding}Micro-averaged $F_1$-scores (averaged over five runs) on English Penn Treebank's Chunking and English CoNLL 2003's NER tasks of models from Section \ref{sec:setup} are plotted (with error bars) against the amount of automatically-labeled data. 0 on the $x$-axis represents models trained with only the clean training set (10k tokens).}   
\end{figure*}
\indent We define the probability of observing a label $z$, which can either be noisy or clean and is, thus, dependent on the label's membership to $C$ or $N$:

{\small
\begin{equation*}
p(z~\shorteq~j|x,w,\theta)~\shorteq~\left\{ \begin{array}{l}
                              \sum^k_{i~\shorteq~1} p(z~\shorteq~j| y~\shorteq~i; \theta)p(y~\shorteq~i|x;w)\\
                              \hfill \quad \text{if}~z \in N\\
                              p(y~\shorteq~j |x;w) \hfill \text{if}~z \in C~\text{i.e.}~z~\shorteq~y 
                             \end{array} \right .
\end{equation*}
}
Using this probability function and $t$ to index training instances, the log-likelihood of the model parameters is defined as

{\small
\begin{align*}
L(w,\theta)&=\sum_{z_t\in C} \log p(z_t|x_t,w)\\[-7pt]
           &+\sum_{z_t \in N} \log( \sum^k_{i~\shorteq~1}(p(z_t|y_t~\shorteq~i;\theta) \cdot p(y_t~\shorteq~i|x_t;w))
\end{align*}
}
\indent As in \citet{7472164} the model parameters are computed using Expectation Maximization. In the E-step, $\theta$ and $w$ are fixed and an estimate $c$ of the true labels $y$ is obtained for the noisy labels $z$:

{\small
\begin{align*}
c_{ti} &~\shorteq~p(y_t~\shorteq~i|x_t,z_t;w,\theta) \\
       &~\shorteq~\frac{p(z_t|y_t~\shorteq~i,\theta)p(y_t~\shorteq~i|x_t;w)}{\sum_jp(z_t|y_t~\shorteq~j;\theta)p(y_t~\shorteq~j|x_t,w)} \hspace{1cm} \text{for } z_t \in N 
\end{align*}
}
\indent Note that the estimate $c$ is calculated only for the noisy labels whereas the clean labels remain unchanged. Similarly, the noise distribution $\theta$ is calculated only for the noisy labels. The initialization of $\theta$ and the $\theta$'s update step in M-step remain the same as in \newcite{7472164}, also shown below.

{\small
 \begin{equation*}
 \theta(i,j) = \frac{\sum_t c_{ti} 1_{\{z_t=j\}}}{\sum_t c_{ti}} \hspace{0.5cm}  i,j \in \{1,...,k\}, z_t \in N
 \end{equation*}
}
During the M-step, the neural network weights $w$ are estimated as well. The loss function, however, changes compared to the original approach \cite{7472164} to (\ref{eqn:loss}) and thus, changing the calculation of the gradient to (\ref{eqn:partial}):

{\small
\begin{equation}
 S(w)=\sum_{z_t\in C} \log p(z_t|x_t,w)
 +  \sum_{z_t\in N} \sum^k_{i~\shorteq~1} c_{ti}\log p(y_t~\shorteq~i|x_t;w) \label{eqn:loss}
\end{equation}
\begin{eqnarray}
 \frac{\partial S}{\partial u_i} ~\shorteq~\sum_{z_t\in C} ( 1_{\{z_t~\shorteq~i\}} - p(z_t|x_t,w))h(x_t) \nonumber\\
 + \sum_{z_t\in N} ( c_{ti} - p(y_t|x_t,w))h(x_t) \label{eqn:partial}
\end{eqnarray}
}
\indent Interestingly, the gradient calculation (\ref{eqn:partial}) is a summation of two parts: one to learn from the clean labels and another to learn from the noisy labels. We refer to this model as the Clean and Noisy Label Neural Network (\textit{CNLNN}).

\section{Training with Noisy Labels in a Multi-Task Setup}
\textit{NLNN} and \textit{CNLNN} form explicit ways of handling noise as the noise distribution is calculated during training. In contrast, we can apply a Deep Multi-Task Learning (MTL) approach \cite{P16-2038}, which, unlike \textit{NLNN} and \textit{CNLNN}, does not estimate the noise directly and thus, is an implicit noise-cleaning approach. The MTL method leverages an auxiliary task that augments the data providing other reliable labels and hence, ignoring noisy labels \cite{DBLP:journals/corr/Ruder17a}. In our experiments, we combine the implicit noise handling of Deep MTL with the explicit noise handling of \textit{NLNN} and \textit{CNLNN}  to complement each other and obtain a more powerful noise handling model than the individual models. Schematic depiction of combining MTL and \textit{CNLNN} is shown in Figure \ref{fig:cnlnn}. MTL and \textit{NLNN} can also be combined in a similar way.

\section{Experimental Setup}
\label{sec:setup} 
We evaluate \textit{CNLNN} and other methods on a Chunking and a Named Entity Recognition (NER) task with $F_1$-score as the metric in each case. For Chunking, we use the same data splits as \cite{P16-2038} based on the English Penn Treebank dataset \cite{Marcus:1993}. For NER, the data splits of the English CoNLL 2003 task are used \cite{tjong2000introduction}. Note that in our NER setup, we evaluate using \textit{BIO-2} labels, so $F_1$-scores reported below might not be comparable to prior work.\\
\indent To mimic a low resource setting, we limit each training set to the first 10k tokens. The development sets are randomly chosen sentences from the original training set restricted to 1k tokens. The test sets remain unchanged. For the rest of the training data, the original labels are removed and the words are automatically labeled using the baseline model (\textit{NN} described below). We add variable amounts of this automatically-annotated data for self-training in our experiments.

\subsection{Models}
\indent We apply the following models to the above two tasks: \textbf{NN} (the simple baseline) is an architecture with bidirectional LSTMs \cite{hochreiter1997lstm}. For Chunking, we use three LSTM layers, for NER five. The \textit{NN} model, only trained on the clean data, is used for automatically labeling the raw data (obtaining the noisy data). \textbf{NLNN} combines the \textit{NN} with the original noise channel \citep{7472164}, training it both on clean and noisy instances. \textbf{CNLNN} is our new approach of modeling noise, treating clean and noisy labels separately (section \ref{sec:cnlnn}).\\
\indent In contrast to the explicit noise handling of \textit{NLNN} and \textit{CNLNN}, we also apply \textbf{MTL} for implicit noise handling.  Here, we use \textit{NN} as the base architecture and POS-tagging as an auxiliary task. We hypothesise that this low-level task helps the model to generalise its representation and that the POS-tags are helpful because e.g. many named entities are proper nouns.  The auxiliary task is trained jointly with the first LSTM layer of \textit{NN} for Chunking and with the second LSTM layer for NER. In our low-resource setting, we use the first 10k tokens of section 0 of Penn Treebank for the auxiliary POS-tagging task for the MTL \cite{P16-2038}. This data is disjunct from the other datasets.\\
\indent Additionally, we combine both the explicit and implicit noise handling. In the low-resource setting, in general, such a combination addresses the data scarcity better than the individual models. \textit{NLNN} and \textit{CNLNN} combinations with MTL are labeled as \textbf{MTL+NLNN} and \textbf{MTL+CNLNN} respectively.

\subsection{Implementation Details}
During training, we minimize the cross entropy loss which sums over the entire sentence. The networks are trained with Stochastic Gradient Descent (SGD). To determine the number of iterations for both the NN model and the EM algorithm we use the development data. All models are trained with word embeddings of dimensionality 64 that are initialized with pre-trained Polygot embeddings \cite{DBLP:conf/conll/Al-RfouPS13}. We add Dropout \cite{JMLR:v15:srivastava14a} with p=0.1 in between the word embedding layer and the LSTM. 

\begin{figure}[t]
	\centering
	\includegraphics[width=\linewidth,
	trim={0 2cm 1.5cm 2.8cm},clip]{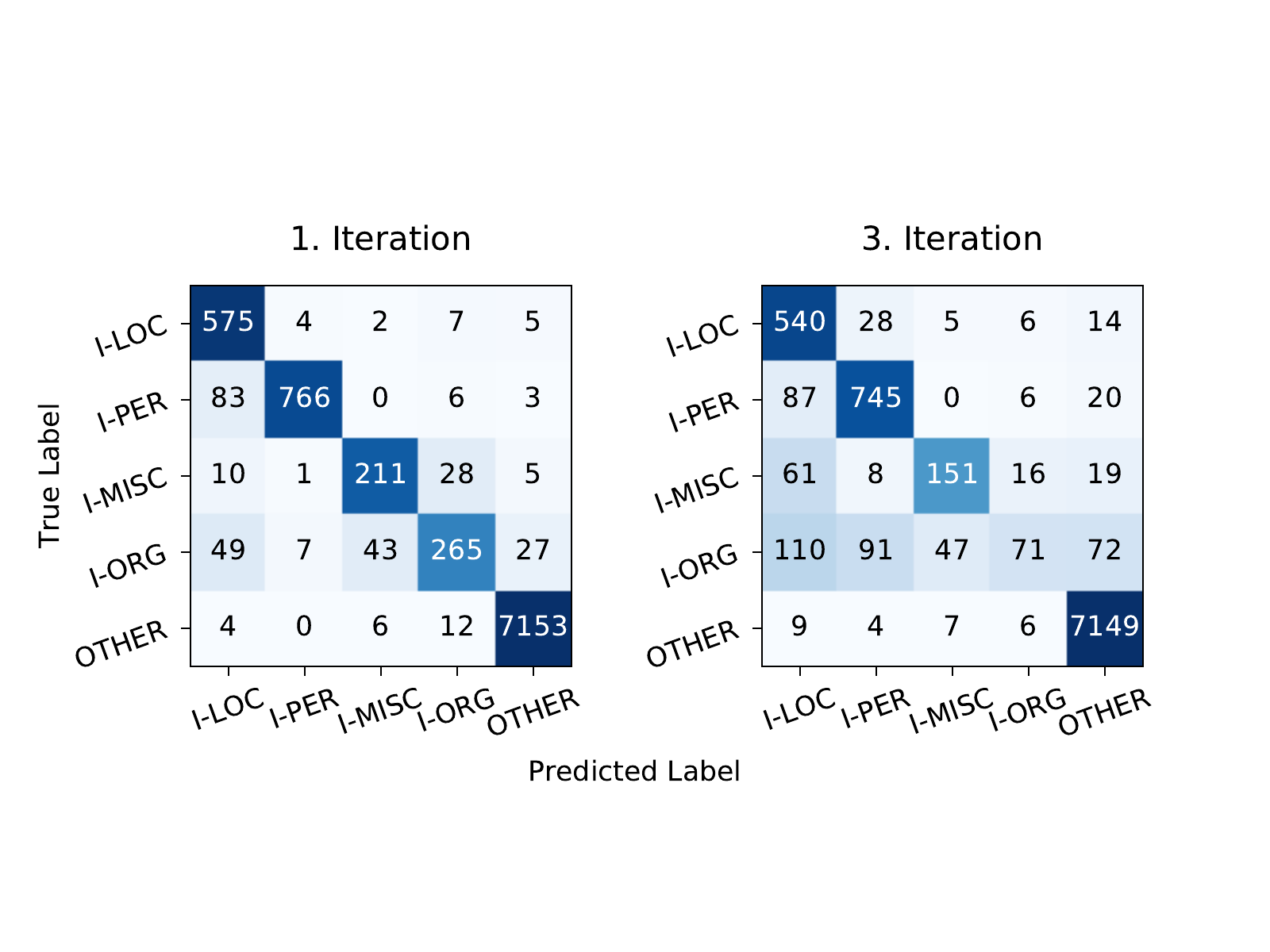}
	\caption{\textit{NLNN} confusion matrices on Chunking's clean training set for 1. and 3. EM iteration. The colors correspond to row-normalized values.}
	\label{fig:em-nlnn}
\end{figure}
\begin{figure}[t]
	\centering
	\includegraphics[width=\linewidth, height=4.5cm,trim={0 0 0 0},clip]{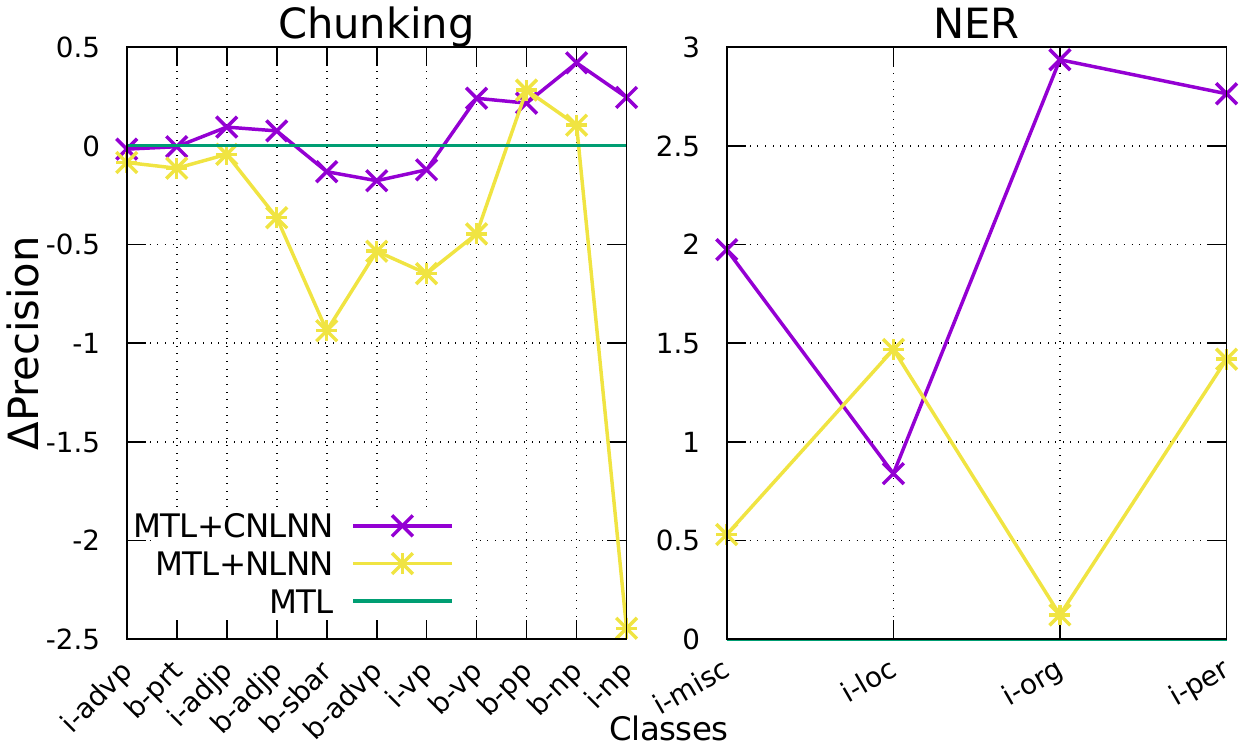}
	\caption{\textit{MTL+CNLNN} vs \textit{MTL+NLNN}: Difference in precision between the combined models and MTL for NER and Chunking test sets with 190K noisy data.}
	\label{fig:cnlnn_vs_nlnn}
\end{figure}
\section{Results}
\label{sec:results}
In Figure \ref{fig:adding}, we present the  $F_1$ scores of the models introduced in the previous section. We perform experiments on Chunking and NER with various amounts of added, automatically-labeled data. In general, adding additional, noisy data tends to improve the performance for all models. This includes the plain \textit{NN}, showing that this model is somewhat robust to noise. Especially for the Chunking task, the possibility for improvement seems limited for \textit{NN} as the performance converges after adding 40k noisy instances. In the Chunking 10k case, the negative effect of the noisy instances results in a score lower than if no data is added.\\
\indent The original \textit{NLNN} model performs similarly to the \textit{NN} model without a noise-handling component. In some cases, the score is even lower. In contrast, \textit{CNLNN} is able to consistently improve over these scores. This demonstrates the importance of our proposed \textit{CNLNN} which treats clean and noisy data separately.\\
\indent \textit{MTL} is able to improve somewhat over \textit{NN} even without adding automatically-annotated data thanks to the auxiliary task. Additionally, \textit{MTL} performs even better when noisy data is added showing its implicit noise handling capabilities. On their own, both \textit{CNLNN} and \textit{MTL} are able to eliminate some of the negative effects of the noisy data and to leverage the additional data effectively.\\
\indent Combining \textit{MTL} with \textit{NLNN} results in small improvements at best and can decrease performance, especially on Chunking. The best results are achieved with our combined \textit{MTL+CNLNN} model as it outperforms all other models. Even when adding 19 times the amount of self-labeled data, the model is still able to cope with the noise and improve the performance.

\section{Analysis}
\label{sec:analysis}
\noindent\ignorespacesafterend \textbf{NLNN vs. CNLNN:} In \textit{NLNN}, we observed that clean training tokens were subverted to become noisy in subsequent EM iterations mostly due to the influence of noisy labels from self-labeled data and this effect leads to \textit{NLNN}'s worse performance. Figure \ref{fig:em-nlnn} presents one such case where the corruption of the confusion matrix from 1. iteration to 3. iteration is displayed. \textit{CNLNN} treats clean and noise data separately and therefore avoids the corruption of clean labels.\\
\noindent\textbf{MTL+CNLNN~vs.~MTL+NLNN:} We noted that \textit{MTL+CNLNN} consistently outperforms \textit{MTL} and \textit{MTL+NLNN}, whereas the \textit{MTL+NLNN} combination can degrade \textit{MTL}'s performance. For nearly all predicted labels the improvements in precision over \textit{MTL} are higher for \textit{MTL+CNLNN} when compared to \textit{MTL+NLNN} (Figure \ref{fig:cnlnn_vs_nlnn}). This shows that \textit{CNLNN} complements \textit{MTL} better than \textit{NLNN}.

\section{Concluding Remarks}
In this paper, we apply self-training to neural networks for Chunking and NER in a low-resource setup. Adding automatically-labeled data, the performance of the classifier can wane or can even decline. We propose to mitigate this effect by applying noisy label handling techniques.\\
\indent However, we found that directly applying an off-the-shelf noise-handling technique as NLNN leads to corruption of the clean training set and worse performance. Thus, we propose the Clean and Noisy Label Neural Network to work separately on the automatically-labeled data. Our model improves the performance faster for a lesser amount of additional data. Moreover, combing the training with auxiliary information can further help handle noise in a complementary fashion.\\
\indent Meanwhile, more complex neural network architectures \cite{Goldberger2017NoiseAdaptation,P17-1040,Veit2017Learning} are available for handling noise and we look forward to working with these to upgrade our approach in the future.

\section{Acknowledgements}
This work has been supported by the German Research Foundation as part of the Research Training Group “Adaptive Preparation of Information from Heterogeneous Sources” (AIPHES) under grant No. GRK 1994/1. We also thank the anonymous reviewers whose comments helped improve this paper.

\bibliography{refs}

\begin{thebibliography}{27}
\expandafter\ifx\csname natexlab\endcsname\relax\def\natexlab#1{#1}\fi

\bibitem[{Al{-}Rfou et~al.(2013)Al{-}Rfou, Perozzi, and
  Skiena}]{DBLP:conf/conll/Al-RfouPS13}
Rami Al{-}Rfou, Bryan Perozzi, and Steven Skiena. 2013.
\newblock Polyglot: Distributed word representations for multilingual {NLP}.
\newblock In \emph{Proceedings of the Seventeenth Conference on Computational
  Natural Language Learning, CoNLL 2013, Sofia, Bulgaria, August 8-9, 2013},
  pages 183--192.

\bibitem[{Bekker and Goldberger(2016)}]{7472164}
Alan~Joseph Bekker and Jacob Goldberger. 2016.
\newblock \href {https://doi.org/10.1109/ICASSP.2016.7472164} {Training deep
  neural-networks based on unreliable labels}.
\newblock In \emph{Proceedings of the 2016 IEEE International Conference on
  Acoustics, Speech and Signal Processing}, pages 2682--2686.

\bibitem[{Charniak(1997)}]{Charniak:1997:SPC:1867406.1867499}
Eugene Charniak. 1997.
\newblock \href {http://dl.acm.org/citation.cfm?id=1867406.1867499}
  {Statistical parsing with a context-free grammar and word statistics}.
\newblock In \emph{Proceedings of the Fourteenth National Conference on
  Artificial Intelligence and Ninth Conference on Innovative Applications of
  Artificial Intelligence}, AAAI'97/IAAI'97, pages 598--603.

\bibitem[{Clark et~al.(2003)Clark, Curran, and
  Osborne}]{Clark:2003:BPT:1119176.1119183}
Stephen Clark, James~R. Curran, and Miles Osborne. 2003.
\newblock Bootstrapping pos taggers using unlabelled data.
\newblock In \emph{Proceedings of the Seventh Conference on Natural Language
  Learning at HLT-NAACL 2003 - Volume 4}, CONLL '03, pages 49--55.

\bibitem[{Dembowski et~al.(2017)Dembowski, Wiegand, and
  Klakow}]{dembowski2017ner}
Julia Dembowski, Michael Wiegand, and Dietrich Klakow. 2017.
\newblock Language independent named entity recognition using distant
  supervision.
\newblock In \emph{Proceedings of Language and Technology Conference (LTC)}.

\bibitem[{Fr{\'e}nay and Verleysen(2014)}]{6685834}
Beno{\^\i}t Fr{\'e}nay and Michel Verleysen. 2014.
\newblock \href {https://doi.org/10.1109/TNNLS.2013.2292894} {Classification in
  the presence of label noise: A survey}.
\newblock \emph{IEEE Transactions on Neural Networks and Learning Systems},
  25(5):845--869.

\bibitem[{Goldberger and Ben-Reuven(2017)}]{Goldberger2017NoiseAdaptation}
Jacob Goldberger and Ehud Ben-Reuven. 2017.
\newblock Training deep neural-networks using a noise adaptation layer.
\newblock In \emph{International Conference on Learning Representations
  (ICLR)}.

\bibitem[{Guan et~al.(2011)Guan, Yuan, Lee, and Lee}]{guan2011Identifying}
Donghai Guan, Weiwei Yuan, Young{-}Koo Lee, and Sungyoung Lee. 2011.
\newblock \href {https://doi.org/10.1007/s10489-010-0225-4} {Identifying
  mislabeled training data with the aid of unlabeled data}.
\newblock \emph{Applied Intelligence}, 35(3):345--358.

\bibitem[{Han et~al.(2018)Han, Yao, Yu, Niu, Xu, Hu, Tsang, and
  Sugiyama}]{Han2018CoteachingRT}
Bo~Han, Quanming Yao, Xingrui Yu, Gang Niu, Miao Xu, Weihua Hu, Ivor~W. Tsang,
  and Masashi Sugiyama. 2018.
\newblock \href
  {http://papers.nips.cc/paper/8072-co-teaching-robust-training-of-deep-neural-networks-with-extremely-noisy-labels}
  {Co-teaching: Robust training of deep neural networks with extremely noisy
  labels}.
\newblock In \emph{Advances in Neural Information Processing Systems 31: Annual
  Conference on Neural Information Processing Systems 2018, NeurIPS 2018, 3-8
  December 2018, Montr{\'{e}}al, Canada.}, pages 8536--8546.

\bibitem[{Hedderich and Klakow(2018)}]{W18-3402}
Michael~A. Hedderich and Dietrich Klakow. 2018.
\newblock \href {http://aclweb.org/anthology/W18-3402} {Training a neural
  network in a low-resource setting on automatically annotated noisy data}.
\newblock In \emph{Proceedings of the Workshop on Deep Learning Approaches for
  Low-Resource NLP}, pages 12--18. Association for Computational Linguistics.

\bibitem[{Hendrycks et~al.(2018)Hendrycks, Mazeika, Wilson, and
  Gimpel}]{Hendrycks2018Trusted}
Dan Hendrycks, Mantas Mazeika, Duncan Wilson, and Kevin Gimpel. 2018.
\newblock \href
  {http://papers.nips.cc/paper/8246-using-trusted-data-to-train-deep-networks-on-labels-corrupted-by-severe-noise}
  {Using trusted data to train deep networks on labels corrupted by severe
  noise}.
\newblock In \emph{Advances in Neural Information Processing Systems 31: Annual
  Conference on Neural Information Processing Systems 2018, NeurIPS 2018, 3-8
  December 2018, Montr{\'{e}}al, Canada.}, pages 10477--10486. Curran
  Associates, Inc.

\bibitem[{Hochreiter and Schmidhuber(1997)}]{hochreiter1997lstm}
Sepp Hochreiter and J{\"u}rgen Schmidhuber. 1997.
\newblock Long short-term memory.
\newblock \emph{Neural computation}, 9(8):1735--1780.

\bibitem[{Luo et~al.(2017)Luo, Feng, Wang, Zhu, Huang, Yan, and
  Zhao}]{P17-1040}
Bingfeng Luo, Yansong Feng, Zheng Wang, Zhanxing Zhu, Songfang Huang, Rui Yan,
  and Dongyan Zhao. 2017.
\newblock Learning with noise: Enhance distantly supervised relation extraction
  with dynamic transition matrix.
\newblock In \emph{Proceedings of the 55th Annual Meeting of the Association
  for Computational Linguistics (Volume 1: Long Papers)}.

\bibitem[{Marcus et~al.(1993)Marcus, Marcinkiewicz, and
  Santorini}]{Marcus:1993}
Mitchell~P. Marcus, Mary~Ann Marcinkiewicz, and Beatrice Santorini. 1993.
\newblock \href {http://dl.acm.org/citation.cfm?id=972470.972475} {Building a
  large annotated corpus of english: The penn treebank}.
\newblock \emph{Computational Linguistics}, 19(2):313--330.

\bibitem[{McClosky et~al.(2006)McClosky, Charniak, and
  Johnson}]{mcclosky2006effective}
David McClosky, Eugene Charniak, and Mark Johnson. 2006.
\newblock Effective self-training for parsing.
\newblock In \emph{Proceedings of the main conference on human language
  technology conference of the North American Chapter of the Association of
  Computational Linguistics}, pages 152--159. Association for Computational
  Linguistics.

\bibitem[{Nettleton et~al.(2010)Nettleton, Orriols-Puig, and
  Fornells}]{Nettleton2010}
David~F. Nettleton, Albert Orriols-Puig, and Albert Fornells. 2010.
\newblock \href {https://doi.org/10.1007/s10462-010-9156-z} {A study of the
  effect of different types of noise on the precision of supervised learning
  techniques}.
\newblock \emph{Artificial Intelligence Review}, 33(4):275--306.

\bibitem[{Pechenizkiy et~al.(2006)Pechenizkiy, Tsymbal, Puuronen, and
  Pechenizkiy}]{Pechenizkiy1647654}
Mykola Pechenizkiy, Alexey Tsymbal, Seppo Puuronen, and Oleksandr Pechenizkiy.
  2006.
\newblock \href {https://doi.org/10.1109/CBMS.2006.65} {Class noise and
  supervised learning in medical domains: The effect of feature extraction}.
\newblock In \emph{19th IEEE Symposium on Computer-Based Medical Systems},
  pages 708--713.

\bibitem[{Petrov et~al.(2010)Petrov, Chang, Ringgaard, and
  Alshawi}]{Petrov:2010:UAD:1870658.1870727}
Slav Petrov, Pi-Chuan Chang, Michael Ringgaard, and Hiyan Alshawi. 2010.
\newblock \href {http://dl.acm.org/citation.cfm?id=1870658.1870727} {Uptraining
  for accurate deterministic question parsing}.
\newblock In \emph{Proceedings of the 2010 Conference on Empirical Methods in
  Natural Language Processing}, EMNLP '10, pages 705--713.

\bibitem[{Ruder(2017)}]{DBLP:journals/corr/Ruder17a}
Sebastian Ruder. 2017.
\newblock \href {http://arxiv.org/abs/1706.05098} {An overview of multi-task
  learning in deep neural networks}.
\newblock \emph{arXiv e-prints}, page arXiv:1706.05098.

\bibitem[{Sagae and Tsujii(2007)}]{sagae2007dependency}
Kenji Sagae and Jun'ichi Tsujii. 2007.
\newblock Dependency parsing and domain adaptation with lr models and parser
  ensembles.
\newblock In \emph{Proceedings of the 2007 Joint Conference on Empirical
  Methods in Natural Language Processing and Computational Natural Language
  Learning}.

\bibitem[{Sang and Buchholz(2000)}]{tjong2000introduction}
Erik F. Tjong~Kim Sang and Sabine Buchholz. 2000.
\newblock Introduction to the conll-2000 shared task: Chunking.
\newblock In \emph{Proceedings of the 2nd workshop on Learning language in
  logic and the 4th conference on Computational natural language
  learning-Volume 7}, pages 127--132. Association for Computational
  Linguistics.

\bibitem[{S{\o}gaard and Goldberg(2016)}]{P16-2038}
Anders S{\o}gaard and Yoav Goldberg. 2016.
\newblock Deep multi-task learning with low level tasks supervised at lower
  layers.
\newblock In \emph{Proceedings of the 54th Annual Meeting of the Association
  for Computational Linguistics (Volume 2: Short Papers)}.

\bibitem[{Srivastava et~al.(2014)Srivastava, Hinton, Krizhevsky, Sutskever, and
  Salakhutdinov}]{JMLR:v15:srivastava14a}
Nitish Srivastava, Geoffrey Hinton, Alex Krizhevsky, Ilya Sutskever, and Ruslan
  Salakhutdinov. 2014.
\newblock \href {http://jmlr.org/papers/v15/srivastava14a.html} {Dropout: A
  simple way to prevent neural networks from overfitting}.
\newblock \emph{Journal of Machine Learning Research}, 15:1929--1958.

\bibitem[{Steedman et~al.(2003)Steedman, Osborne, Sarkar, Clark, Hwa,
  Hockenmaier, Ruhlen, Baker, and Crim}]{Steedman:2003:BSP:1067807.1067851}
Mark Steedman, Miles Osborne, Anoop Sarkar, Stephen Clark, Rebecca Hwa, Julia
  Hockenmaier, Paul Ruhlen, Steven Baker, and Jeremiah Crim. 2003.
\newblock Bootstrapping statistical parsers from small datasets.
\newblock In \emph{Proceedings of the Tenth Conference on European Chapter of
  the Association for Computational Linguistics - Volume 1}, EACL '03, pages
  331--338.

\bibitem[{Sukhbaatar et~al.(2015)Sukhbaatar, Bruna, Paluri, Bourdev, and
  Fergus}]{Sukhbaatar2015Learning}
Sainbayar Sukhbaatar, Joan Bruna, Manohar Paluri, Lubomir Bourdev, and Rob
  Fergus. 2015.
\newblock Learning from noisy labels with deep neural networks.
\newblock In \emph{ICLR Workshop track}.

\bibitem[{Veit et~al.(2017)Veit, Alldrin, Chechik, Krasin, Gupta, and
  Belongie}]{Veit2017Learning}
Andreas Veit, Neil Alldrin, Gal Chechik, Ivan Krasin, Abhinav Gupta, and Serge
  Belongie. 2017.
\newblock \href
  {http://openaccess.thecvf.com/content_cvpr_2017/papers/Veit_Learning_From_Noisy_CVPR_2017_paper.pdf}
  {Learning from noisy large-scale datasets with minimal supervision}.
\newblock In \emph{Proceedings of the IEEE Conference on Computer Vision and
  Pattern Recognition}, pages 839--847.

\bibitem[{Zhang and Sabuncu(2018)}]{Zhang2018GeneralizedCE}
Zhilu Zhang and Mert~R. Sabuncu. 2018.
\newblock \href
  {http://papers.nips.cc/paper/8094-generalized-cross-entropy-loss-for-training-deep-neural-networks-with-noisy-labels}
  {Generalized cross entropy loss for training deep neural networks with noisy
  labels}.
\newblock In \emph{Advances in Neural Information Processing Systems 31: Annual
  Conference on Neural Information Processing Systems 2018, NeurIPS 2018, 3-8
  December 2018, Montr{\'{e}}al, Canada.}, pages 8792--8802.

\end{thebibliography}
\bibliographystyle{acl_natbib}

\end{document}